\documentclass[a4paper,fleqn]{cas-dc}
\usepackage[numbers]{natbib}

\usepackage{tcolorbox}
\usepackage{soul}
\usepackage{xcolor}
\usepackage{soul}

\soulregister\cite7

\def\tsc#1{\csdef{#1}{\textsc{\lowercase{#1}}\xspace}}
\tsc{WGM}
\tsc{QE}
\tsc{EP}
\tsc{PMS}
\tsc{BEC}
\tsc{DE}
\RenewDocumentCommand \printorcid {} {}
\begin{document}
\let\WriteBookmarks\relax
\def\floatpagepagefraction{1}
\def\textpagefraction{.001}
\shorttitle{InterRSGaussian}
\shortauthors{Z.Qi, Z.Zou, Z.Shi.}

\title [mode = title]{InterRSGaussian: Interactive Multi-view Instance Segmentation for 3D Remote Sensing Scenes} 
\tnotemark[1]
\author[1]{Zipeng Qi}
\credit{Methodology, Writing -- Original Draft, Validation, Visualization}

\author[1]{Zhengxia Zou}
\credit{Methodology, Writing -- Review and Editing}

\author[1,2]{Zhenwei Shi}
\cormark[1]
\credit{Writing -- Review and Editing, Supervision}

\cortext[cor1]{Corresponding author}

\affiliation[1]{
  organization={School of Astronautics, Beihang University},
  city={Beijing},
  postcode={100191},
  country={China}
}

\affiliation[2]{
  organization={Key Laboratory of Spacecraft Design Optimization and Dynamic Simulation Technologies, Ministry of Education, Beihang University},
  city={Beijing},
  postcode={100191},
  country={China}
}
\tnotetext[1]{This manuscript has been accepted for publication in Fundamental Research. The final published version will be available on the journal website.}

\begin{abstract}
Remote sensing image segmentation plays a critical role in various applications. Interactive instance segmentation, a specialized task within image segmentation, aims to generate precise segmentation results for target objects based on user input prompts. However, previous studies on interactive instance segmentation primarily focus on single-view images, limiting their ability to comprehensively understand complex 3D remote sensing scenes. In this study, we extend interactive instance segmentation from a single-view to a multi-view framework, which can also be easily converted to a 3D segmentation format, thus achieving more accurate and comprehensive 3D remote sensing scene perception across multiple views. First, we construct a multi-view interactive instance segmentation dataset for remote sensing, comprising five distinct scenes, each captured from about one hundred different views. Second, we propose InterRSGaussian, a novel instance segmentation method based on 3D Gaussians. InterRSGaussian enables users to provide point clicks on target objects from any arbitrary view and generates view-consistent instance segmentation results across all input views without requiring manual annotations. The resulting multi-view segmentation can further be projected onto 3D point clouds for 3D segmentation. To improve training efficiency and segmentation performance, we introduce a prior distillation network for initialization. Overall, this study extends traditional 2D remote sensing scene perception into a 3D multi-view framework, enabling a more comprehensive understanding of target objects in complex environments with user interactivity.
\end{abstract}


\maketitle

\section{Introduction}
Remote sensing image segmentation~\cite{kotaridis2021remote,yuan2021review, li2024review, wang2020comprehensive} is a fundamental perception task in remote sensing image processing. It assigns a category label to each pixel in the input image, enabling the extraction of critical information such as object location, boundaries, and spatial distribution. Image segmentation tasks can be broadly classified into semantic segmentation~\cite{diakogiannis2020resunet,li2021multiattention, BIE2025} and instance segmentation~\cite{ su2019object,hafiz2020survey,xu2021improved}. Semantic segmentation assigns a semantic label to each pixel, grouping regions with the same category. Furthermore, instance segmentation not only differentiates between semantic categories but also identifies and distinguishes individual objects within the same category, assigning each instance a unique identify ID. This distinction makes instance segmentation particularly valuable for applications such as object detection~\cite{cao2020d2det,RU2023}, object counting~\cite{cholakkal2019object}, and scene understanding~\cite{du2024two}, where precise object delineation is crucial for accurate analysis and decision-making.

Interactive instance segmentation builds upon traditional instance segmentation by incorporating user prompts—such as point clicks, bounding boxes, or text descriptions—to guide the segmentation process. This user-driven paradigm enables more precise and flexible object extraction while enhancing user control through selective perception and analysis of objects of interest. Furthermore, segmentation results can be dynamically refined in real time based on user input, thereby improving both flexibility and practical usability. By integrating user feedback directly into the segmentation loop, this approach supports iterative refinement for more accurate and customized outcomes. 
Despite significant advancements in interactive instance segmentation~\cite{xu2022land, chen2024rsprompter, li2025insight, zhang2024faster}, previous research has primarily focused on processing single-view remote sensing images. However, these methods remain inherently constrained to 2D perception, limiting their ability to fully capture the spatial structure and understanding within complex remote sensing objects and scenes. 

With the rapid advancement of 3D scene reconstruction technology~\cite{mildenhall2021nerf, qi20223d, li20213d, kerbl20233d, MENG20251169}, the demand for enhanced 3D perception and understanding has grown significantly. This need is especially critical in remote sensing applications, where 3D perception enables a more comprehensive analysis of ground objects from multiple views~\cite{qi2023implicit, 10642693, DBLP:journals/tgrs/QiCZZS25, DBLP:journals/corr/abs-2309-07509}. Leveraging 3D scene understanding allows for more accurate and efficient execution of key tasks such as urban planning, infrastructure monitoring, and environmental assessment, ultimately facilitating more informed decision-making. This study extends traditional 2D remote sensing scene perception into a 3D multi-view framework that can be easily converted to other 3D formats, such as point cloud segmentation, enabling a deeper and more comprehensive understanding of objects and complex environments.

\begin{figure*}
	\centering
	\includegraphics[width=1.0\textwidth]{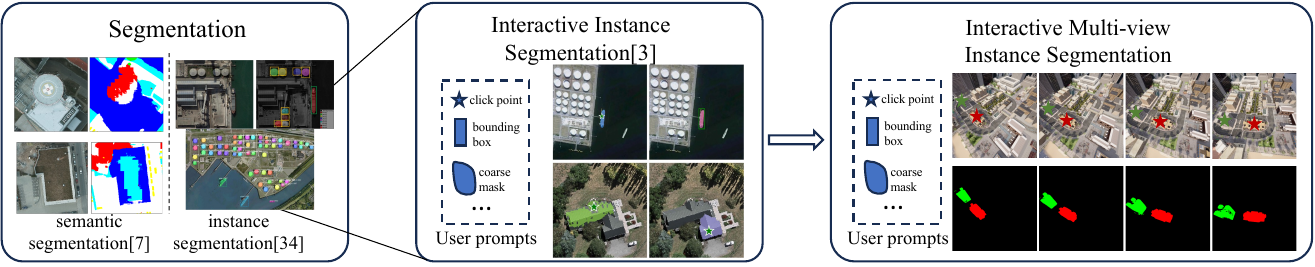}
	\caption{\textbf{The multi-view interactive instance segmentation for 3D remote sensing scenes}.Instance segmentation is a specialized task within segmentation. Interactive instance segmentation enables the segmentation of user-specified prompt objects. This study extends the previous single-view interactive instance segmentation to a multi-view framework, enhancing the ability of comprehensively understanding objects and scenes.}
	\label{FIG:1}
\end{figure*}

Unlike previous studies, this study takes multi-view images of 3D remote sensing scenes along with user prompts (e.g., point clicks) as input and generates instance segmentation results for the objects of interest in each view, as illustrated in Figure~\ref{FIG:1}. Therefore, the primary research objective of this study is not only to achieve accurate segmentation results but also to ensure the consistency of the segmentation outcomes across different views for the prompt objects. 
Maintaining strong view consistency is essential for ensuring the reliability of downstream processes, such as transforming multi-view segmentation results into point cloud (3D) segmentation or other forms of representation. Beyond perception outcomes, view consistency is also critical for accurately analyzing the 3D structure of target objects in 3D remote sensing scenes.

To ensure view consistency in segmentation results, this study extracts 3D spatial representations from input multi-view remote sensing images. Various 3D spatial representations have been employed for modeling 3D remote sensing scenes, including point clouds~\cite{zhang2017advances,guo2020deep,liu2019deep}, TSDFs (Truncated Signed Distance Fields)\cite{qi20223d,fehr2017tsdf}, latent neural representations\cite{mildenhall2021nerf,shi2022deep}, and 3D Gaussians~\cite{kerbl20233d, chen2024survey,fei20243d}, all of which have become prominent research directions. Among these, 3D Gaussians offer an effective balance between accuracy and efficiency. Accordingly, this study adopts 3D Gaussians as the core 3D spatial representation for 3D remote sensing scenes and proposes a novel method for multi-view instance segmentation based on this representation. In particular, the method aims to eliminate the need for manual annotations during training or optimization, thereby significantly reducing labor costs and improving practical scalability.

To achieve segmentation without manual labeling, this study leverages the prior knowledge embedded in foundation models~\cite{kirillov2023segment, zhang2022dino}. Foundation models have garnered significant attention in artificial intelligence, particularly within deep learning, due to their ability to generalize across tasks. Typically, a foundation model is pre-trained on large-scale datasets~\cite{deng2009imagenet, lin2014microsoft}, and through extensive exposure to diverse data, it develops robust, general-purpose feature extraction capabilities. This adaptability enables these models to be effectively applied to a wide range of downstream tasks without requiring task-specific training from scratch. For instance, models from the SAM series~\cite{kirillov2023segment,zhang2023survey,zhang2023towards, zhang2024segment} can automatically or interactively segment a wide variety of scenes, producing 2D segmentation results that demonstrate strong generalization ability. Similarly, models from the DINO series~\cite{zhang2022dino,li2023mask} exhibit universal visual feature extraction capabilities, efficiently learning useful visual representations through unsupervised learning, which makes them highly versatile for various downstream applications. The large model priors not only provide high-quality initial segmentation but also offer robust visual features for initialization. 

Overall, this study focuses on the task of interactive image segmentation, which holds significant value, and extends traditional 2D perception to multi-view stereo perception. A novel method for multi-view interactive instance segmentation in remote sensing is proposed, incorporating 3D spatial information representations of the scene and leveraging foundational model priors to enable segmentation without the need for manual labels. Furthermore, this study explores the use of foundation model priors as initialization to accelerate training/optimization convergence and improve segmentation performance. Additionally, an interactive multi-view instance segmentation dataset specifically designed for 3D remote sensing scenes is constructed.

\section{Material}
Datasets for interactive multi-view instance segmentation are scarce in the field of remote sensing image processing. To address this problem, we first construct a related dataset in this study. 
\begin{figure*}
	\centering
	\includegraphics[width=1.0\textwidth]{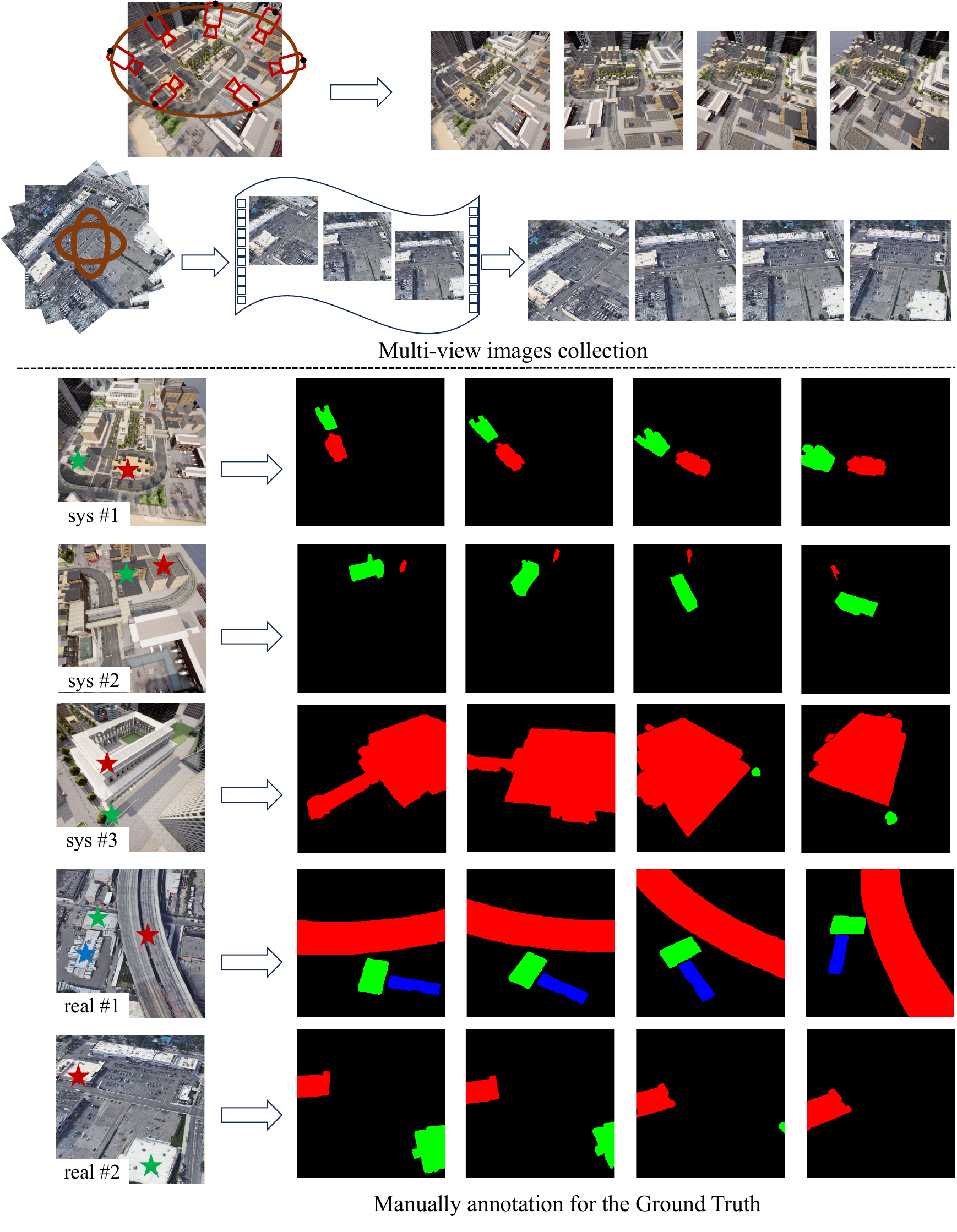}
	\caption{\textbf{In this study, we construct a multi-view interactive instance segmentation dataset for remote sensing}. This dataset consists of five scenes including three simulated scenes and two real-world scenes.}
	\label{FIG:2}
\end{figure*}

The constructed dataset comprises five 3D remote sensing scenes, including both simulated and real-world data. The simulated data are collected using the CARLA platform~\cite{dosovitskiy2017carla}. By configuring the position and pitch angle of the simulated camera in a 3D space, remote sensing images are generated from varying heights and viewpoints. The real-world data collection is conducted using the Google Maps platform and is intended solely for academic research. First, we select the target scene and adjust the pixel resolution to ensure clear image details. Then, we utilize the 3D view mode of Google Maps, enabling free rotation to observe the target area from multiple perspectives. Using the perspective rotation and screen recording functions provided by Google Maps, we manually adjust the viewpoint to perform a comprehensive scan around the target scene, capturing multi-view video data. Finally, key frames are selected from the recorded video.

After collecting multi-view images for each scene, we select 2-3 objects per scene to simulate user selection. The segmentation ground truth for each object in each view is then manually annotated. The construction process is illustrated in Figure~\ref{FIG:2} and the details of the constructed dataset is shown in Table~\ref{TAB:dataset}.
\begin{table}[width=.9\linewidth,cols=4,pos=h]
\caption{The details of constructed dataset.}
\label{TAB:dataset}
\begin{tabular*}{\tblwidth}{@{} LLLL@{} }
\toprule
Scene & Type & View num. & Object num.\\
\midrule
sys \#1 & simulation data & 100 & 2 \\
sys \#2 & simulation data & 80 & 2 \\
sys \#3 & simulation data & 85 & 2 \\
real \#1 & real-world data & 100 & 3 \\
real \#2 & real-world data & 100 & 2 \\
\bottomrule
\end{tabular*}
\end{table}

The dataset constructed in this study considers objects of varying sizes, ranging from large structures such as buildings and bridges to smaller elements like pots and bushes. It also incorporates challenging conditions, including object occlusions and scenarios where objects are not visible from one view but become apparent from another. These factors introduce additional complexity, thereby enhancing the dataset's realism and its utility for robust evaluation.

\section{Method}
\begin{figure*}
	\centering
	\includegraphics[width=.9\textwidth]{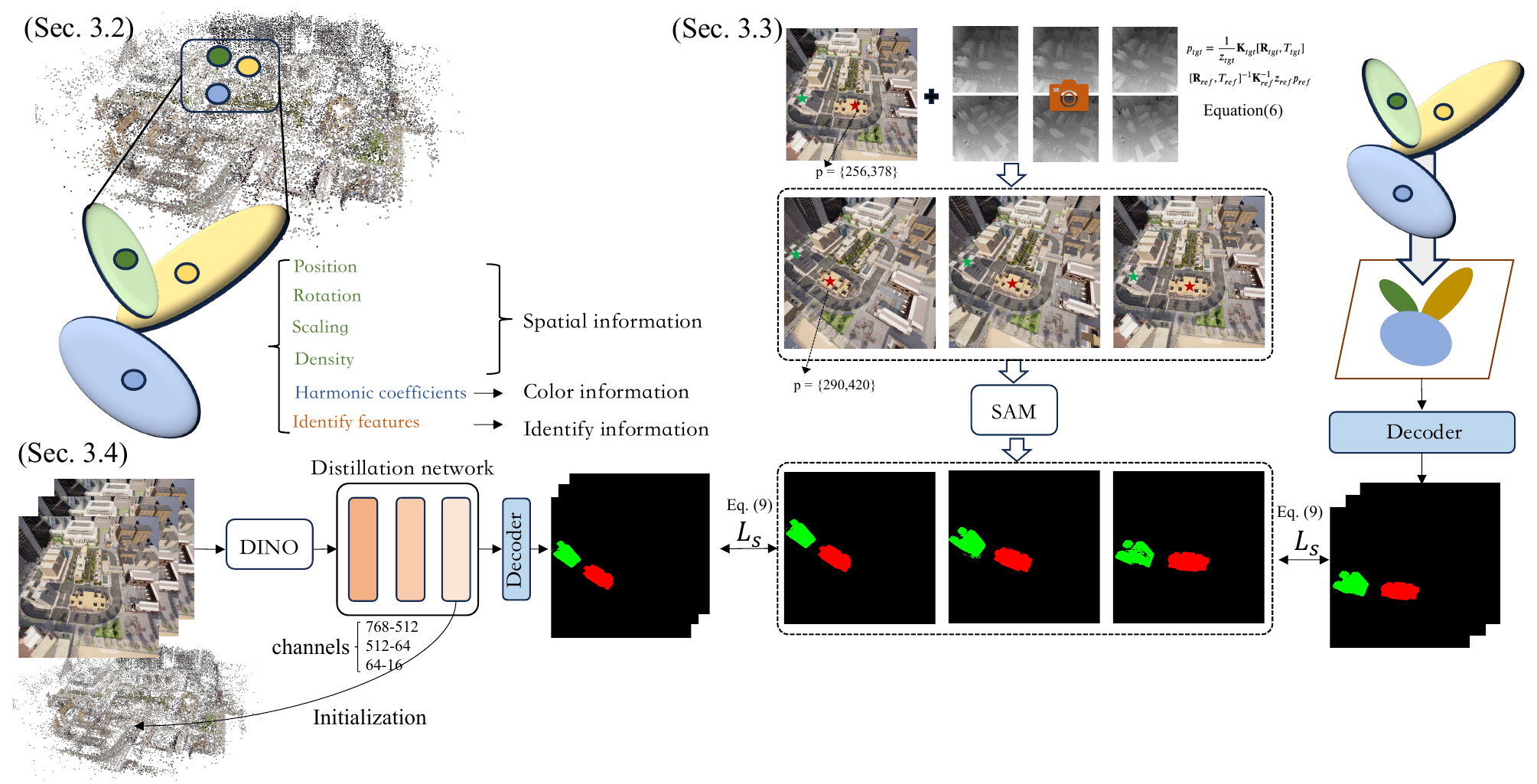}
	\caption{\textbf{In this study, we propose an unsupervised multi-view instance segmentation method based on 3D Gaussians, termed InterRSGaussian}. InterRSGaussian consists of 3D spatial representation(Sec. 3.2), transferring 2D prompts into multi-view(Sec. 3.3), and DINO distillation for initialization(Sec. 3.4).}
	\label{FIG:3}
\end{figure*}

This study proposes an multi-view instance segmentation method based on 3D Gaussians, termed InterRSGaussian. InterRSGaussian leverages pre-reconstructed 3D Gaussians or reconstructs the 3D spatial representation of the input multi-view images, providing a structured foundation for instance segmentation across different views. InterRSGaussian enables users to prompt target objects using point clicks in arbitrary view image and subsequently generates multi-view instance segmentation results for each view. The proposed method integrates 3D Gaussians for 3D spatial representation extraction, identity feature learning and pseudo-label generation for transferring 2D prompts into multi-view segmentation, DINO distillation for initialization to accelerate training convergence. The overview of the method is shown in Figure~\ref{FIG:3}.

\subsection{Task description}
This study focuses on multi-view instance segmentation based on user input prompts for 3D remote sensing scenes. The input consists of \( n \) view images: \([I_1, \dots, I_n]\) and \( m \) prompt objects, each with \( l \) point clicks. If there are no pre-reconstructed 3D Gaussians, we need to pre-reconstruct the 3D Gaussians using the method described in Section~\ref{3D Gaussians}. The output is the instance segmentation for each prompt object across all input view images, represented as:\(\underbrace{[O^1_1, \dots, O^1_n]}_{\text{object}_1:O_1}, \dots, \underbrace{[O^m_1, \dots, O^m_n]}_{\text{object}_m:O_m}\). And, the multi-view segmentation result can be easily converted to 3D format, e.g, point cloud segmentation.

\subsection{3D Gaussians}
\label{3D Gaussians}
The 3D Gaussians form the foundational concept of InterRSGaussian for extracting 3D spatial representation from multi-view images. Thus, we first provide a brief introduction to 3D Gaussians' optimization process.

The 3D Gaussians initialize the scene's point cloud using COLMAP (\hyperlink{}{https://colmap.github.io/}), assigning a 3D Gaussian, $g$, to each point. Each Gaussian is characterized by a mean \(\mu^{3D} \in \mathbb{R}^3\) representing the point's position and a covariance matrix \(\Sigma^{3D} = RSS^T R^T\) defining its 3D spatial distribution. The scaling matrix \(S \in \mathbb{R}^3\) and rotation matrix \(R \in \mathbb{R}^{3 \times 3}\) determine the Gaussian transformation. Additionally, the density parameter \(\sigma \in \mathbb{R}^1\) represents opacity. The above make up the 3D spatial information. For the color information, spherical harmonic coefficients \(H \in \mathbb{R}^{27}\) encode view-dependent appearance. The color vector \(c \in \mathbb{R}^3\) is computed using camera parameters and \(H\) based on the rendering angle.

During rendering, each 3D Gaussian is projected onto the image plane of the selected view, forming a 2D Gaussian with mean \(\mu^{2D} \in \mathbb{R}^2\) and covariance matrix \(\Sigma_i^{2D} \in \mathbb{R}^{2 \times 2}\). The mean \(\mu^{2D}\) is calculated by the coordinate transformation from the world to the image plane  based on \(\mu^{3D}\). The covariance matrix \(\Sigma_i^{2D}\) is computed using the following equation:
\begin{equation}
\Sigma_i^{2D} = J W \Sigma_i^{3D} W^T J^T,
\end{equation}
where \(J\) is the Jacobian matrix for the 3D-to-2D projection, and $W$ transforms world coordinates to camera coordinates and consists of both camera intrinsics and extrinsics. For a pixel in \(p \in \mathbb{R}^2\) position from a selected view image, \(n\) Gaussians are selected whose covariance matrix influence range covers \(p\) and are closest to the image plane at point \(p\). These Gaussians are then synthesized using alpha blending:
\begin{equation}
C^{p} = \sum_{i \in n} c_i \alpha'_i \prod_{j=1}^{i-1} \Big(1 - \alpha'_j\Big),
\end{equation}
where \(\alpha'_i\) is the transparency:
\begin{equation}
\alpha'_i = \sigma_i \exp\left(-\frac{1}{2}(p - \mu_i^{2D})^T (\Sigma_i^{2D})^{-1}(p - \mu_i^{2D})\right).
\end{equation}
The colors of other pixels in the image are obtained in parallel using the same process. Next, the optimization is performed using \(\mathcal{L}_{1}\) and \(\mathcal{L}_{D-SSIM}\) losses~\cite{kerbl20233d}:
\begin{align}
&\mathcal{L}_{1} = |C - \hat{C}|,\\
&\mathcal{L}_{\text{LD-SSIM}} = 1 - \frac{\sum_{i=1}^{N} w_i \cdot \text{SSIM}(C_i, \hat{C}_i)}{\sum_{i=1}^{N} w_i},
\end{align}

where \(\hat{C}\) and \(C\) are the ground truth and whole rendered image of the selected view, respectively. The parameters of each Gaussian $g$ are updated/optimized by the loss according to the gradient backpropagation. The whole optimization process involves repeatedly changing views to optimize the parameters of all Gaussians.

The optimization refines the Gaussian parameters, including point positions and spherical harmonic coefficients, which represent the 3D spatial structure and appearance of the scenes. Accurate color rendering is essential for scene reconstruction, as it enhances realism and preserves 3D spatial structure. The Gaussian mean and variance jointly define the 3D spatial representation of the scene, which are frozen in the later process. 

All attributes are initialized from the SfM sparse point cloud: positions are directly assigned; colors use RGB-to-SH conversion for the DC component with higher-order SH set to zero; scaling is initialized via KNN distances followed by log transformation; rotation starts as a unit quaternion; opacity is initialized to 0.1 (via inverse sigmoid)

\subsection{Transfer 2D prompts into multi-view segmentation}
Given the per-reconstruction 3D Gaussian representation of the scenes, InterRSGaussian allows the user to prompt the target object by point-clicking in any arbitrary view image. InterRSGaussian then transfers the 2D prompts into multi-view segmentation.

\textbf{Pseudo label generation.} To eliminate the need for manual annotations during training/optimization, we employ the SAM model to generate pseudo-instance labels for each view based on the user's point-click input. The first step is to propagate the point prompts across all views. Specifically, the user provides a few point-click positions for each target object on a selected view image, referred to as the reference view. To facilitate homogeneous representation, we augment each reference point-click coordinate $(u,v)_{ref}$ by adding a latitude component with a value of 1, forming \( p_{ref} = (u,v,1)_{ref} \). We then project these reference coordinates onto the corresponding positions in the remaining views, referred to as the target views, denoted as \( p_{tgt} = (u,v,1)_{tgt} \). The transformation process for each pixel follows:
\begin{equation}
p_{tgt} = \frac{1}{z_{tgt}}\textbf{K}_{tgt} [\textbf{R}_{tgt}, T_{tgt}] [\textbf{R}_{ref}, T_{ref}]^{-1} \textbf{K}_{ref}^{-1} z_{ref} p_{ref},
\end{equation}
where \( z_{ref} \) represents the depth value of \( (u,v)_{ref} \) and \( z_{tgt} \) represents the depth value of \( (u,v)_{tgt} \), both of which indicate the respective distances from the camera in their respective coordinate systems. These depth value for a pixel is rendered from the 3D Gaussians as following:
\begin{equation}
    z = \sum_{i \in n} d_i \alpha'_i \prod_{j=1}^{i-1} \Big(1 - \alpha'_j\Big),
\end{equation}
where, $d_i$ is the depth distance of $i$-th Gaussian $g$, assigned to the pixel, to the image plane.
The parameters \( \textbf{K}_{tgt}\), \(\textbf{R}_{tgt}\), \(T_{tgt}\), \(\textbf{K}_{ref}\), \(\textbf{R}_{ref}\), \(T_{ref}\) correspond to the intrinsic and extrinsic camera parameters of the reference and target views, respectively, and can be pre-estimated using COLMAP. The second step involves generating pseudo-instance labels by feeding the propagated point prompts into the SAM model. We process the SAM output to one-hot labels, where the label for $i$-th view is denoted as $\hat{O}_i\in \mathcal{R}^{H\times W \times m}$. Here, $W$ and $H$ are the image size parameters. Since the point prompts have been transferred across multiple views, SAM can segment the corresponding object in each view independently. Although this process may introduce some errors due to inconsistencies in the propagated prompts, the subsequent identity feature learning will help refine and correct these errors, ensuring more robust and coherent multi-view instance segmentation.

\textbf{Identity feature learning.} We assign each Gaussian $g$ in the pre-reconstruction 3D Gaussians an identity feature, \(f \in \mathbb{R}^{16}\). Similar to color rendering, we select a view and render the identity features onto the image plane at the pixel position \(p\) using alpha blending:
\begin{equation}
F^{p} = \sum_{i \in n} f_i \alpha'_i \prod_{j=1}^{i-1} \Big(1 - \alpha'_j\Big),
\end{equation}
where \(n\) and \(\alpha'_i\) are the same as in the color rendering process. The identity features for other pixels in the image plane are also obtained in parallel using the same method. We define \(F\) as the complete set of identity features for the entire image. Then, a one-layer decoder is introduced to transform the 2D identity features into segmentation results:
\begin{equation}
O = \text{Decoder}(F).
\end{equation}
The identity feature optimization is performed using a cross-entropy loss function as follows:
\begin{equation}
\label{eq:cross_entropy}
\mathcal{L}_s = - \frac{1}{H W} \sum_{i=1}^{H} \sum_{j=1}^{W} \sum_{d=1}^{m} \hat{y}_{ijd} \log y_{ijd},
\end{equation}
where, \( \hat{y}_{ijd} \) is the pseudo instance label of the selected view at pixel position \( (i,j) \), \( y_{ijd} \) is the predicted instance label, and \( m \) represents the number of prompted objects. The identity features are updated/optimized by the $\mathcal{L}_s$ according to the gradient backpropagation. The entire optimization process involves iteratively switching views to refine the identity features of all Gaussians, ensuring alignment with the pseudo instance labels. Once the optimization is complete,  we can render instance segmentation for each view. Thanks to the uniform representation of identity information using 3D Gaussians, the segmentation results maintain great view consistency.

\subsection{DINO distillation for initialization.} 
We set random initialization as the default setting for identity features. To accelerate convergence, we instead use distilled DINO features for initializing these identity features. The DINO features have dimensions \(\frac{H}{14} \times \frac{W}{14} \times 768\), where \(H\) and \(W\) denote the height and width of the input image, and 768 is the number of feature channels. Since the vanilla DINO features have a channel mismatch with the required identity features, they cannot be used directly. To resolve this, we introduce a distillation network composed of three convolutional layers. This network takes as input the upsampled, pre-extracted DINO features from each view image. We then employ the same decoder used in identity feature learning to generate the segmentation results. The pseudo instance labels serve as supervision, and we calculate the loss using Equation~\ref{eq:cross_entropy} to optimize the parameters of the distillation network. Figure~\ref{FIG:3} details the channel configuration of the distillation network, where the final layer is set to output 16 channels to align with the identity features.

Then, we generate the coarse 3D mask for the pre-trained 3D Gaussians. Specifically, we project the world coordinates, \( p_w = \{x, y, z\} \), of each point in the 3D Gaussians onto each view image plane:
\begin{equation}
    (u,v,k)^T_i = \mathbf{K}_i [\mathbf{R}_i, T_i]\, p_w,
\end{equation}
where \( \left( \frac{u}{k}, \frac{v}{k} \right)^T_i \) represents the pixel coordinates in the \( i \)-th view image, and \( \mathbf{K}_i \), \( \mathbf{R}_i \), and \( T_i \) denote the corresponding camera parameters for the \( i \)-th view. If \( \left( \frac{u}{k}, \frac{v}{k} \right)^T_i \) falls within the image boundaries, the corresponding object ID from the pseudo-label is assigned to that point. Ultimately, the object ID of each point is determined by the most frequently assigned object ID across all views, thereby forming the coarse 3D mask.

Finally, for each prompt object and the background, we randomly select a view and obtain the distilled features by sampling one distilled DINO feature according to the pseudo-labels. We then assign the distilled DINO features corresponding to each prompt object and the background to the associated points in the 3D Gaussians based on the 3D mask, serving as the initialization features. The experiments imply that this distilled initialization accelerates convergence efficiency and improves segmentation performance. Additionally, the distillation network is a three-layer CNN with channel transformations: 768→512, 512→64, and 64→16, which are shown in Figure~\ref{FIG:3}.

\section{Results}
\begin{figure*}[!h]
	\centering
    \includegraphics[width=0.89\textwidth]{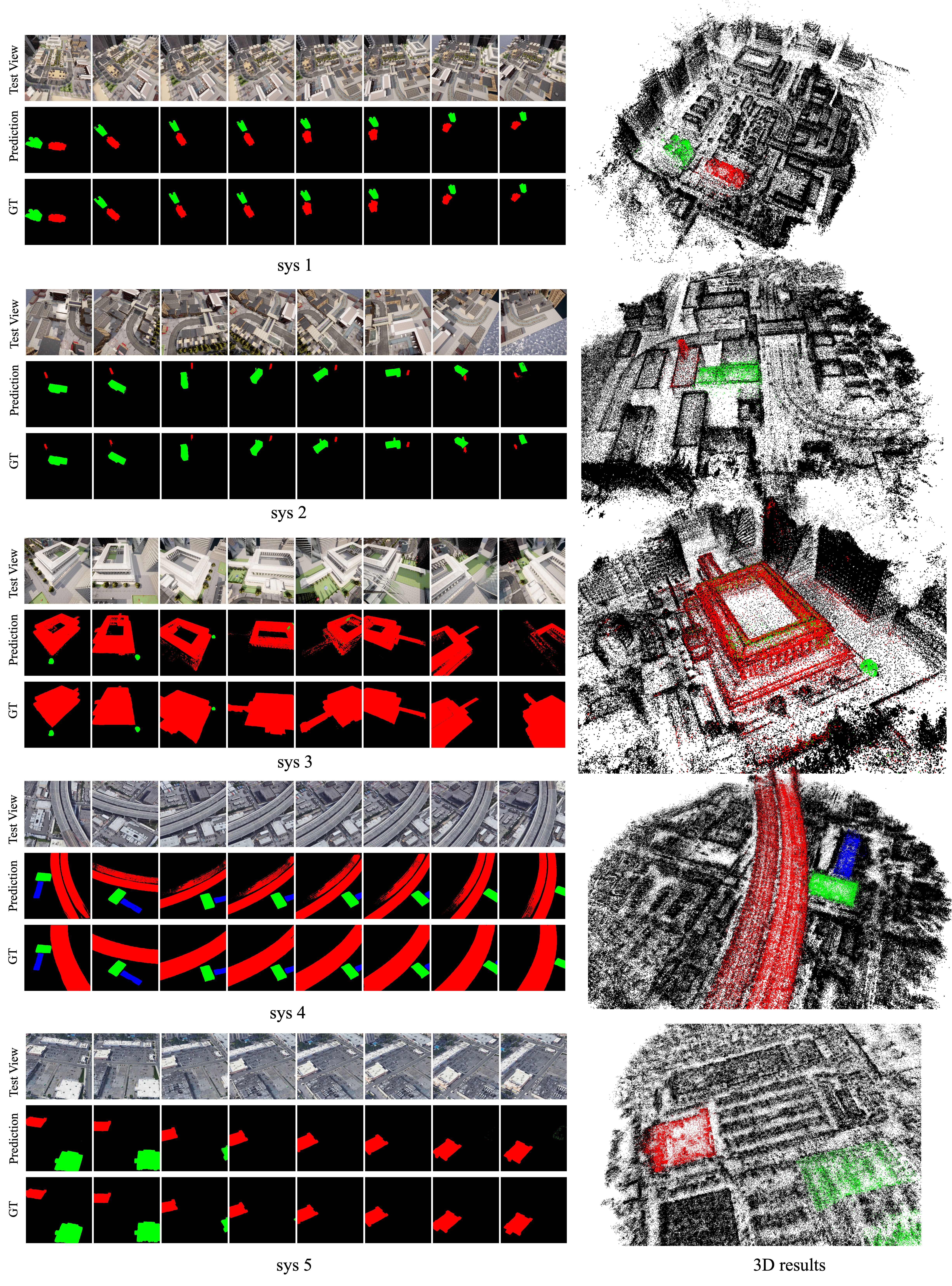}
	\caption{\textbf{Qualitative results}. This study presents the 2D and 3D segmentation results for each scene.}
	\label{FIG:4}
\end{figure*}
Our experiments were performed on a platform featuring an RTX 3090 GPU and an Intel Core i9 CPU, utilizing the constructed dataset. We also analyze the runtime performance on this platform. The average total runtime for segmenting objects across ~100 views is 760 s. In addition to the GPU usage of the SAM and DINO networks, the spatial representation extraction and identity feature optimization for 3D Gaussian require an average additional 10–12 GB of GPU memory. In the training process, for 3D spatial representation extraction, positions follow an exponential decay learning rate, while colors, scaling, rotation, and opacity use constant rates. The default learning rate is set to 1e-5. For identity feature learning, semantic features are initialized from 2D DINO features and optimized at a learning rate of 1e-5 with exponential decay, while all other parameters are frozen. The distillation network is initialized using Kaiming initialization and optimized with a learning rate of 1e-5.
The Table~\ref{TAB:2} shows the quantitative results and Figure~\ref{FIG:4} shows the qualitative results.
\subsection{Quantitative results}
We compare the mIoU metrics of the pseudo labels from SAM, the results from the compared method, Gaussian Grouping~\cite{gaussian_grouping}, and our method. We retrain the Gaussian Grouping on our dataset and also use the SAM results as training labels, the same as our method. We ensure that the channel number of the identity features in both Gaussian Grouping and our method is set to 16 for a fair comparison. The optimization iteration in DINO distillation is set to 5,000 and in identify feature optimization is set to 1,000 as our default setting. The number of prompt click $l$ is set to 5. Table~\ref{TAB:2} shows the results.
\begin{table}[width=1.0\linewidth,cols=6,pos=h]
\caption{The quantitative results.}
\label{TAB:2}
\begin{tabular*}{\tblwidth}{@{} LLLLLL@{} }
\toprule
Method & sys \#1 & sys \#2 & sys \#3 &real \#1&real \#2\\
\midrule
SAM & 94.77 & 83.86 & 50.62&89.88& 96.79 \\
Gaussian Grouping-1k & 96.19 & 89.92 & 71.13&92.24&96.76 \\
Gaussian Grouping-10k & 96.36 & 89.94 & 59.43&92.49&90.58 \\
Ours-1k & \textbf{96.52} & 90.10 & \textbf{76.62}&\textbf{93.18}& \textbf{96.81} \\
Ours-10k & 96.40 & \textbf{90.14} & 56.67& 91.43 & 90.83  \\
\bottomrule
\end{tabular*}
\end{table}

We observe that all methods perform well in terms of quantitative results, with SAM performing the worst. The main reasons are as follows:(1) SAM has not been trained on a large number of remote sensing images, which makes it challenging to process such type images.(2) SAM processes each view image independently and does not exploit the 3D spatial relationship between different views. (3) SAM has difficulty coping with situations where objects disappear and suddenly reappear in certain views, and struggles to distinguish multiple objects that are occluded in certain views.

Although SAM cannot handle every view perfectly and struggles with occlusions and sudden appearances of objects in some views, it still provides high-quality pseudo labels in most views. Based on these pseudo-labels, Gaussian Grouping and our method produce higher-quality results. By using the 3D spatial representation of 3D Gaussians, the optimization of the identity feature of each point is informed by valid pseudo label information from all views. Therefore, even if the pseudo label quality of SAM is low in some views, it can be compensated by other high-quality views.

In the comparison between Gaussian Grouping and our method, we show the results of two optimization scenarios. One involves optimizing 1k times, and the other involves optimizing 10k times. The suffix *-1k represents the setting of optimizing 1k times, while *-10k represents the setting of optimizing 10k times. 

The results show that with the same number of optimizations, the method proposed in this study outperforms Gaussian Grouping in all scenes. Especially in sys \#3, under the optimization setting of 1k, our method outperforms Gaussian Grouping by more than 5 percentage points. The main reason is that sys \#3 contains large-scale objects and occlusions between objects, leading to low-quality pseudo-labels. However, the method proposed in this study uses the DINO distillation prior, and the initialized identity features themselves have the ability to distinguish between objects, allowing it to better handle low-quality pseudo-labels.

By comparing the results of different optimization iterations, we found that the benefit of increasing the number of optimizations is not obvious (This will be further discussed in Sec. 5). This is reflected in both Gaussian Grouping and the proposed method. For example, in sys \#1, more optimizations only provide a 0.17 improvement for Gaussian Grouping, which has decreased in most scenarios for the proposed method. However, increasing the number of optimizations will greatly increase the training/optimization time. Therefore, under the setting of 1k training/optimization times, the proposed method achieves the best results with less training time.

\subsection{Qualitative results}
We show the qualitative results of the proposed method in Figure~\ref{FIG:4} with randomly selected 10 views for each scene.

First, we can observe that the segmentation of prompt objects for all scenes is strongly consistent across views. This is due to the unified identity representation in 3D spatial form within 3D Gaussians.

Second, we can observe that the segmentation results of sys \#1 and \#2 are the most complete and accurate. This is mainly because the prompt objects in these two scenes are not blocked and do not disappear or appear from certain views, so the pseudo-label quality of SAM is relatively high. In contrast, the green prompt building in real \#2 disappears from certain views, resulting in a decrease in the pseudo-label quality of SAM. Therefore, using this as supervision leads to a decrease in segmentation quality.

Finally, the segmentation of objects of different sizes is compared, and it is found that the proposed method has better performance for small and medium-sized objects. This is because large objects require more and denser prompts, and the 5 prompt points set in this study cannot guarantee that the edges of large objects can be correctly segmented by SAM in every view. For example, the segmentation accuracy of the edges of the buildings in sys \#3 and the bridges in real \#1 is relatively poor.

Additionally, we discuss other output formats of our method. In addition to the multi-view segmentation format, we can directly input the identity features of each point in the 3D Gaussians into the decoder to obtain the segmentation. This process outputs the segmentation results in point cloud format, as shown in the right part of Figure~\ref{FIG:4}. Since ground truth for point cloud segmentation is unavailable, we cannot perform a quantitative analysis of these results. However, the qualitative results are highly promising. The segmentation results in point cloud format enable us to further analyze the 3D spatial distribution of each prompt object. The results with this format can be particularly valuable in several applications, such as environmental monitoring, urban planning and development, geospatial data analysis, and more.

\begin{figure*}[!t]
	\centering
	\includegraphics[width=.9\textwidth]{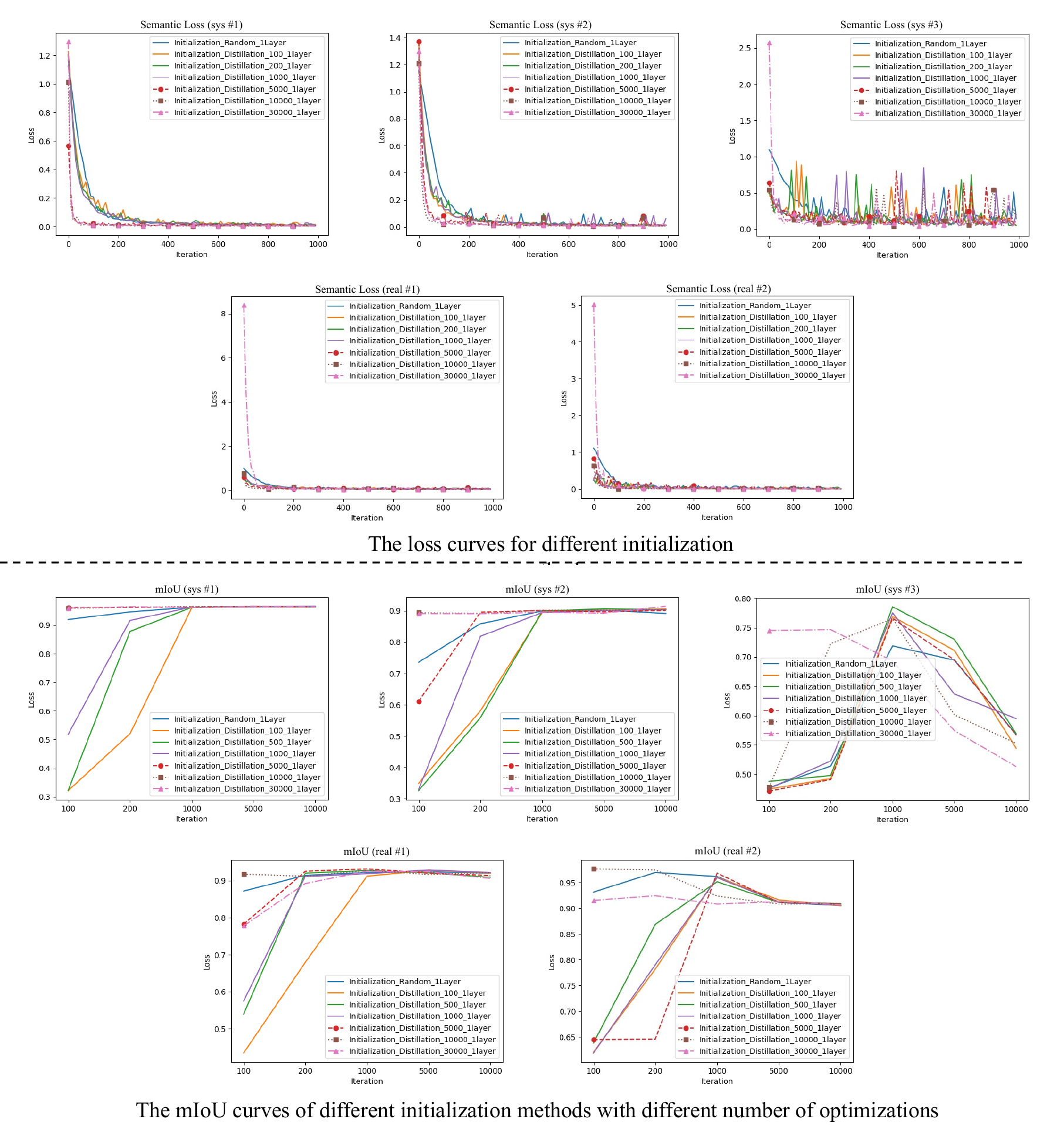}
	\caption{\textbf{Loss curves and mIoU values.}This study discuss the impact of different initialization methods on convergence speed and training/optimization iterations on segmentation accuracy.}
	\label{FIG:5}
\end{figure*}
\section{Discussion}
In order to fully assess the effectiveness of the proposed method, this study explores the number of convolutional layers in the decoder, the necessity of DINO distillation, the number of training iterations in the distillation process, and the number of identity feature optimizations. The related results are shown in Table~\ref{TAB:3} and Figure~\ref{FIG:5}.

\begin{table}[width=1.0\linewidth,cols=6,pos=h]
\caption{The discussion of the number of decoder layers and DINO distillation. The results in the upper part are obtained without DINO distillation, while the results in the lower part correspond to DINO distillation with a 1-layer decoder. And, we also compare the distillation iteration number.}
\label{TAB:3}
\begin{tabular*}{\tblwidth}{@{} LLLLLL@{} }
\toprule
Method & sys \#1 & sys \#2 & sys \#3 &real \#1&real \#2\\
\midrule
1-Layer & 96.21 & 89.96 & 62.95&92.24&96.17 \\
2-Layer & 96.22 & 90.01 & 73.55&92.10&93.89 \\
3-Layer & 96.23 & 90.13 & 75.00&92.39&92.40 \\
4-Layer & 96.05 & 64.79 & 48.41&92.44&91.85 \\
\hline
Distillation-100 & 96.10 & 89.47 & 76.91&91.25&96.01 \\
Distillation-500 & 96.30 & 89.87 & 78.59&92.72&95.14 \\
Distillation-1k & 96.30 & 89.45 & 77.55&91.91& 95.92\\
Distillation-5k & \textbf{96.52} & 90.10 & \textbf{76.62}&\textbf{93.18}&\textbf{96.81} \\
Distillation-1w & 96.54 & \textbf{90.19} & 76.56&92.76& 92.36\\
Distillation-3w & 96.20 & 89.55 & 69.31&92.55&90.80 \\
\bottomrule
\end{tabular*}
\end{table}

First, we gradually increase the number of convolutional layers in the decoder from the default value of one to four. As the number of layers increases, the optimization time significantly increases. However, the mIoU metric results do not show a clear difference, and in some cases, even a decrease, as shown in the upper part of Table~\ref{TAB:3}. This is because, while a more complex network architecture improves representation ability, it also amplifies sensitivity to the accumulated errors in pseudo-labels. 
The results show that a simpler decoder architecture can achieve good performance and effectiveness.

Next, we discuss the necessity of DINO distillation and the number of optimization iterations in the distillation process. The results in the lower half of Table~\ref{TAB:3} are based on the default 1-layer decoder with the addition of distilled DINO prior as the initialization of the identity feature. The suffix *-number represents the setting of training iteration for distillation. The results show that excessive distillation optimization can cause the initialization features to overly align with the DINO feature space, which increases the difficulty in identity feature learning. On the other hand, insufficient distillation optimization may fail to learn effective priors. These observations are indicated in Table~\ref{TAB:3}. Thus, we set the optimization iteration number to 5,000 as the default setting to extract a suitable DINO prior. This setting shows great improvement. In particular, for complex sys \#3 with large-scale buildings and occlusion, the distillation operation has the greatest effect on improving segmentation accuracy. This indicates that the distilled features themselves have a certain degree of differentiation for different objects, making the segmentation task simpler and improving the segmentation performance.

Then, we compared the convergence performance of different initialization methods during the training/optimization process of identity features. From the upper part of Figure~\ref{FIG:5}, it can be seen that in each scene, when the number of distillation training iteration is greater than 1,000, the convergence speed is significantly faster compared to the random initialization. Among them, the setting of 5,000 distillation training iteration has the fastest convergence speed and the lowest semantic loss at the beginning of training/optimization. The above results show that the DINO prior distillation feature is better than the randomly initialized feature, and choosing the right distillation iteration is also very necessary.

Finally, we discuss the change in segmentation accuracy as the training/optimization iterations increase under different initialization methods. From the results shown in the lower part of Figure~\ref{FIG:5}, we can observe that regardless of the initialization method, the segmentation accuracy becomes optimal or stabilizes within 1,000 training/optimization iterations in almost all scenes. Compared with random initialization, distillation initialization can achieve stability faster in the accuracy change curve after more than 1,000 distillation training iterations. For example, the Distillation-10k and Distillation-30k settings reach stability after only 100 training/optimization iterations in sys \#1 and \#2. However, in a comprehensive comparison of all scenarios, Distillation-5k generally achieves the best segmentation accuracy, with a fast convergence speed. Therefore, this study uses Distillation-5k as the default setting for initialization, and the training/optimization iteration number for identity feature learning is set to 1,000.

\subsection{Comparison with Related Methods}
Most 3D Gaussian Splatting studies summarized in~\cite{wu2024recent} focus on reconstruction, editing, and related downstream applications. InterRSGaussian extends this representation to interactive instance segmentation by associating identity features with 3D Gaussians in remote sensing. The unified 3D representation enables consistent in segmentation results across views.

The method in~\cite{yang2017view} suggests informative views and requires users to draw strokes in multiple views. In contrast, InterRSGaussian requires only point prompts in a single view and automatically propagates the results to other views, providing a more convenient interaction process. However, it does not explicitly select the optimal view and may be sensitive to occlusion or depth errors in the selected view.

The methods in~\cite{sun2023semantic,tian2026bridging} perform supervised instance segmentation directly on 3D point clouds by exploiting semantic and geometric cues. Although they provide explicit 3D reasoning, they require semantic and instance annotations for training. InterRSGaussian instead uses SAM pseudo-labels and DINO features, reducing annotation costs while supporting category-independent selection of user-specified objects. However, it does not explicitly model part hierarchies or jointly reason about geometry and semantics, and remains sensitive to pseudo-label noise.

\section{Limitation}
We acknowledge several limitations. First, directly applying general-purpose foundation models to remote sensing images may be suboptimal because of domain gaps. Parameter-efficient fine-tuning (PEFT) methods, such as adapters, prompt tuning, and low-rank adaptation, could improve the SAM pseudo-labels and DINO initialization features with limited trainable parameters~\cite{yin2025remote}. We will leave PEFT-based adaptation for future work. Second, relying on SAM-generated pseudo-labels introduces noise, and handling extremely thin or ambiguous boundaries remains challenging. To address this issue, we plan to develop an end-to-end framework that jointly optimizes 3D structure, color, and identity features. Third, the number of scenes in our dataset is limited, although we have ensured coverage of major urban object types with diverse sizes and textures. We will continue expanding the dataset to strengthen our findings.

\section{Conclusion}
This study extends interactive instance segmentation from the conventional single-view setting to a multi-view setting. The proposed InterRSGaussian utilizes 3D Gaussian to generate view-consistent segmentation results from user-provided point prompts. We also construct a multi-view interactive instance segmentation dataset containing five 3D scenes, each captured from approximately 100 views. To accelerate convergence and improve segmentation performance, we introduce a prior distillation network to initialize the identity features. Experimental results demonstrate the effectiveness of the proposed method. By extending interactive perception from 2D images to multi-view 3D scenes, our framework enables more consistent and comprehensive analysis of user-specified objects in complex remote sensing environments.
\printcredits

\section*{Declaration of competing interest}

The authors declare that they have no conflicts of interest in this
work.

\section*{Acknowledgment}
The work was supported by the National Natural Science Foundation of China under the Grants 62125102, U24B20177, and 62471014.

\bibliographystyle{cas-model2-names}

\bibliography{cas-refs}

\end{document}